\documentclass[runningheads]{llncs}

\usepackage{eccv}
\usepackage[width=122mm,left=12mm,paperwidth=146mm,height=193mm,top=12mm,paperheight=217mm]{geometry}

\usepackage{eccvabbrv}

\usepackage{graphicx}
\usepackage{booktabs}
\usepackage{caption}
\usepackage{graphicx}
\usepackage{xcolor}
\usepackage{booktabs}
\usepackage{amsmath}     
\usepackage{array}       
\usepackage{colortbl}    

\usepackage[accsupp]{axessibility}  

\newcommand{\methodname}{RefineAnything}

\usepackage[breaklinks,colorlinks,citecolor=eccvblue]{hyperref}

\usepackage{orcidlink}

\begin{document}

\title{RefineAnything: Multimodal Region-Specific Refinement for Perfect Local Details} 

\titlerunning{RefineAnything}

\author{Dewei Zhou\inst{1} \and
You Li\inst{1} \and
Zongxin Yang\inst{2} \and
Yi Yang\inst{1}}

\authorrunning{D.~Zhou et al.}

\institute{RELER, CCAI, Zhejiang University, \email{\{zdw1999, uli2000, yangyics\}@zju.edu.cn} \and
DBMI, HMS, Harvard University, \email{Zongxin\_Yang@hms.harvard.edu}}

\maketitle

\begin{center}
\begin{minipage}{0.9\linewidth}
\centering  
\includegraphics[width=\linewidth]{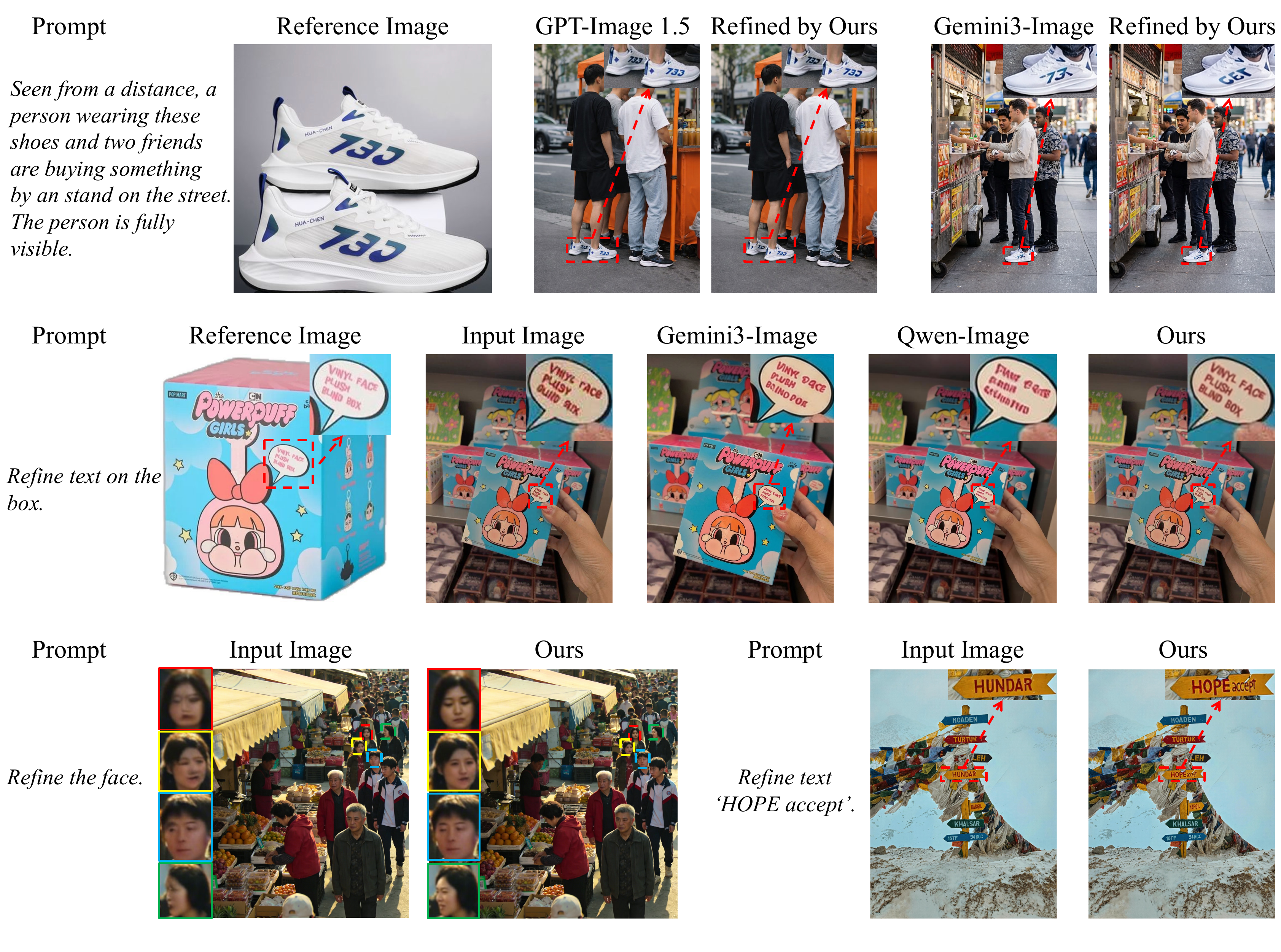}  
\captionof{figure}{
\textbf{RefineAnything} restores fine-grained details (e.g., text, logos, and faces) in user-specified regions (indicated by the bounding boxes) for both reference-based and reference-free inputs, keeping the background unchanged.
}
\label{fig:teaser}
\end{minipage}
\end{center}
\label{fig:teaser}

\begin{abstract}
 We introduce \emph{region-specific image refinement} as a dedicated problem setting: given an input image and a user-specified region (e.g., a scribble mask or a bounding box), the goal is to restore fine-grained details while keeping all non-edited pixels \emph{strictly unchanged}. Despite rapid progress in image generation, modern models still frequently suffer from \emph{local detail collapse} (e.g., distorted text, logos, and thin structures). Existing instruction-driven editing models emphasize coarse-grained semantic edits and often either overlook subtle local defects or inadvertently change the background, especially when the region of interest occupies only a small portion of a fixed-resolution input.
We present \textsc{RefineAnything}, a multimodal diffusion-based refinement model that supports both reference-based and reference-free refinement. Building on a counter-intuitive observation that crop-and-resize can substantially improve local reconstruction under a fixed VAE input resolution, we propose \emph{Focus-and-Refine}, a region-focused refinement-and-paste-back strategy that improves refinement effectiveness and efficiency by reallocating the resolution budget to the target region, while a blended-mask paste-back guarantees strict background preservation. We further introduce a boundary-aware \emph{Boundary Consistency Loss} to reduce seam artifacts and improve paste-back naturalness.
To support this new setting, we construct \textbf{Refine-30K} (20K reference-based and 10K reference-free samples) and introduce \textbf{RefineEval}, a benchmark that evaluates both edited-region fidelity and background consistency. On RefineEval, \textsc{RefineAnything} achieves strong improvements over competitive baselines and near-perfect background preservation, establishing a practical solution for high-precision local refinement. Project Page: \url{https://limuloo.github.io/RefineAnything/}.

\end{abstract}

\section{Introduction}
Image generation has advanced rapidly, and modern models offer substantially improved controllability~\cite{seedream2025seedream,hidreami1technicalreport,zhang2025eligen,zhou2025dreamrenderer,ipcir,liu2025step1x,zhang2025enabling,gpt-4o,bagel,wu2025qwen,zhou2024migc,zhou20243dis,chen2025ragd,du2025textcrafter,chen2025dip,li2024anysynth,li2025controlnet,li2024controlnet++,zhang2024creatilayout,li2025seg2any,zhang2025creatidesign,shi2025consistcompose,zhou2025bidedpo,xu2025contextgen,lu2023tf,lu2024mace,lu2024robust,lu2025does,zhou2025dragflow,zhao2023wavelet,zhao2025zero,zhao2025ultrahr,zhao2024toward,zhao2024learning,zhao2026luve,li2026foleydirector}. Yet a practical failure mode still frequently blocks real-world deployment: \emph{local detail collapse}. As shown in Fig.~\ref{fig:teaser}, fine-grained elements such as printed text, logos, and thin structures are often distorted or inconsistent, even when the global composition is plausible. This issue is particularly damaging in high-stakes applications where small details carry key information, such as e-commerce product images and advertisements, retail signage and packaging, or UI/infographics, where a single wrong character or broken stroke can undermine trust and usability.

This motivates \emph{region-specific image refinement} as a dedicated problem setting: given an input image and a user-specified region, the goal is to \emph{improve local details} while keeping the rest of the image \emph{strictly unchanged}.

In this setting, a natural first attempt is to use today’s instruction-driven editing models to ``fix'' local defects with prompts. However, existing paradigms are not well-suited to refinement, as shown in Fig.~\ref{fig:teaser} and Fig.~\ref{fig:qualitative_result}, mainly due to three issues: (1) \textbf{weak region controllability}---it is difficult to precisely specify \emph{where} to refine; (2) \textbf{poor micro-detail recovery}---subtle defects (e.g., broken text strokes) are often left unresolved; and (3) \textbf{background drift}---non-target regions may change unintentionally. In practice, users require a refinement tool that is simultaneously \emph{region-accurate}, \emph{detail-effective}, and \emph{background-preserving}.

\begin{figure*}[t]
  \centering
   \includegraphics[width=\linewidth]{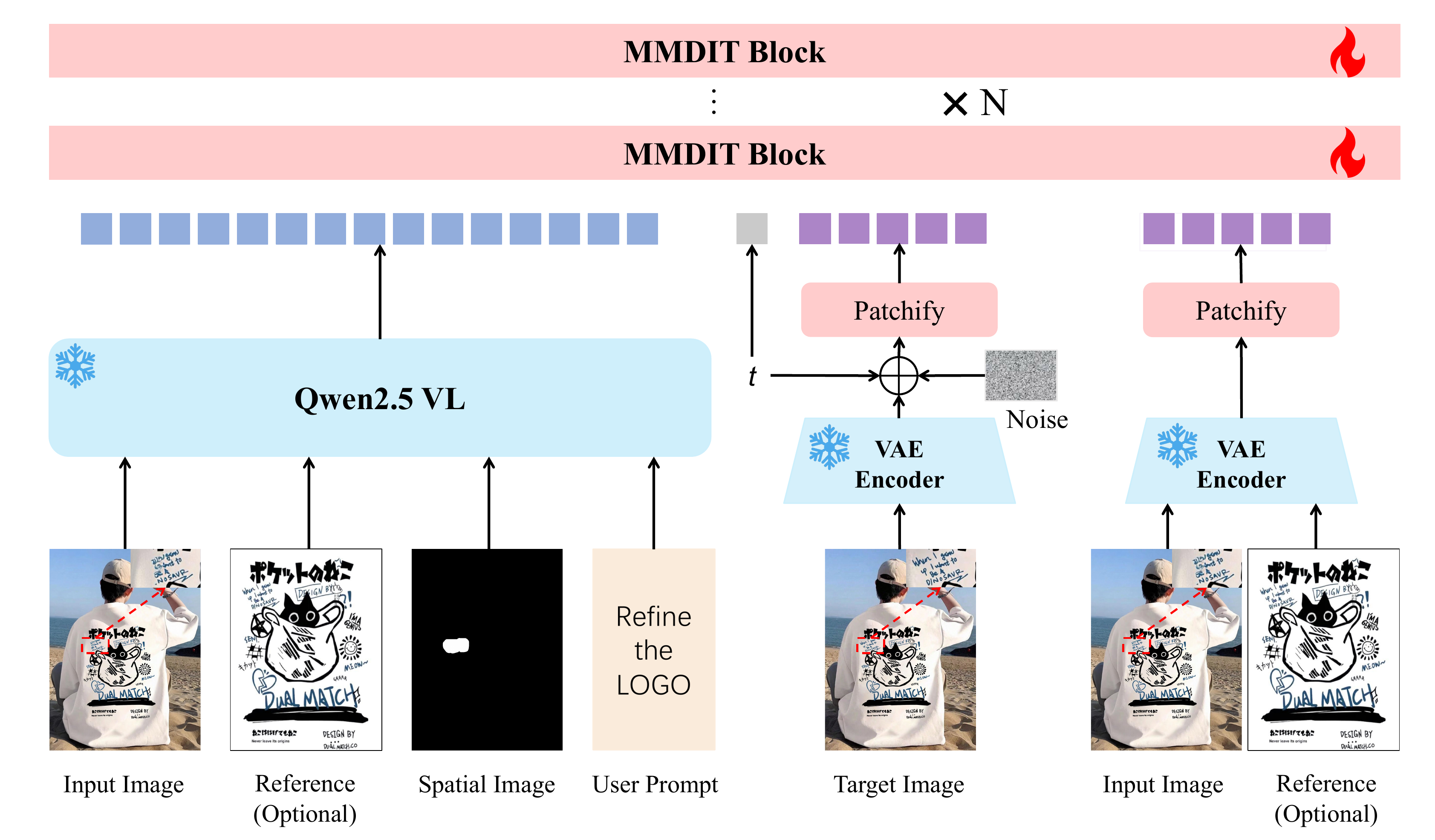}
\vspace{-5mm}
\caption{\textbf{Architecture of RefineAnything.} Given an input image and an optional reference image, the user specifies an edit region via a scribble mask; the images, region cue, and text instruction are encoded by a frozen Qwen2.5-VL encoder into multimodal conditioning tokens. Conditioned on these tokens, a diffusion backbone built from MMDiT blocks (trainable, e.g., via LoRA~\cite{hu2022lora,xu2024ctrlora}) denoises a VAE latent from timestep $t$ to produce the locally refined result.}
\label{fig:arch}
\vspace{-3mm}

\end{figure*}

To achieve \textbf{region controllability}, we propose \textbf{RefineAnything} (Fig.~\ref{fig:arch}), a region-aware refinement model that builds on recent multimodal editing models~\cite{wu2025qwen} and fine-tunes them with explicit region cues. RefineAnything injects region cues (scribbles or bounding boxes) into the model’s conditioning, enabling user-specified refinement in both reference-based and reference-free settings.
%
Nevertheless, \textbf{micro-detail recovery} remains challenging when the target region is very small (see Fig.~\ref{fig:vs_focus_and_refine}), since most modern diffusion models generate in the VAE latent space and decoding from latents inevitably incurs information loss~\cite{vae,stablediffusion,wu2025qwen}; this loss becomes more pronounced when the region itself contains only a limited amount of \emph{effective} pixel information. This motivates a counter-intuitive yet impactful observation (Fig.~\ref{fig:why_focus}): simply cropping a small target region and upsampling it to the same resolution as the full image---upsampling does not increase the amount of effective pixel information---can yield substantially better VAE reconstruction \emph{within the region} than reconstructing the full image. Building on this, we introduce \emph{Focus-and-Refine} (Fig.~\ref{fig:focus_and_paste}): we refine the focused crop and paste it back with a blended mask, improving refinement effectiveness and efficiency.
Focus-and-Refine also naturally enforces \textbf{background preservation}: the blended-mask paste-back guarantees strict background consistency by construction. To further improve paste-back naturalness, we propose a \emph{Boundary Consistency Loss} that strengthens training supervision near the edit boundary to reduce seam artifacts. 

To support training and evaluation at scale, we build \textbf{Refine-30K}, a dataset of 30K samples (20K reference-based and 10K reference-free) constructed with VLM grounding, SAM-based segmentation, and controlled inpainting degradations while explicitly preserving the background. We also introduce \textbf{RefineEval}, a benchmark that evaluates both edited-region fidelity and background preservation in reference-based and reference-free settings.

Extensive experiments on RefineEval show that \textbf{RefineAnything} consistently outperforms the strongest baselines: it improves region fidelity with lower MSE/LPIPS~\cite{pydiff} reconstruction errors (0.020/0.155 vs.\ 0.040/0.264), and strengthens semantic alignment with higher DINO~\cite{zhang2022dino,oquab2023dinov2}/CLIP~\cite{clip} similarities and SSIM~\cite{wang2004imagessim} scores (0.793/0.885/0.591 vs.\ 0.675/0.807/0.436). Meanwhile, it achieves near-perfect background consistency with lower $\mathrm{MSE}_{bg}$/$\mathrm{LPIPS}_{bg}$ errors and higher $\mathrm{SSIM}_{bg}$ scores (0.000/0.000/0.9997 vs.\ 0.011/0.019/0.9660).

In summary, our contributions are three-fold:
\begin{itemize}
    \item We formulate \emph{region-specific image refinement} as a new setting and present \textsc{RefineAnything}, a practical system that improves local details while keeping non-edited regions strictly unchanged.
    \item We propose \emph{Focus-and-Refine} and a boundary-aware \emph{Boundary Consistency Loss} to enable high-quality refinement with seamless paste-back.
    \item We construct \textbf{Refine-30K} and \textbf{RefineEval} to support training and evaluation in both reference-based and reference-free settings, and demonstrate strong improvements in refinement quality, semantic alignment, and background consistency.
\end{itemize}

\section{Related Work}

\noindent\textbf{Image Generation Models.}
Image generation has progressed rapidly, delivering high-fidelity images with stronger controllability and instruction following. Modern models largely build upon diffusion models~\cite{ddpm}. In particular, the Stable Diffusion family (SD1.5~\cite{stablediffusion}, SDXL~\cite{sdxl}) popularizes latent diffusion, where a variational autoencoder (VAE)~\cite{vae} maps images into a compact latent space for denoising, significantly accelerating training and sampling; many subsequent models adopt this VAE-based latent framework. Building on this foundation, the community has moved from UNet backbones to better-scaling Diffusion Transformers, such as Hunyuan-DiT~\cite{li2024hunyuandit}, PixArt~\cite{pixart}, SD3~\cite{sd3}, and FLUX~\cite{flux}. More recently, multimodal generators (e.g., Qwen-Image~\cite{wu2025qwen} and Flux Klein~\cite{flux}) incorporate VLM encoders (e.g., Qwen2.5-VL~\cite{Qwen2.5-VL}) to jointly interpret text and images, broadening real-world applications. Nevertheless, even state-of-the-art models still struggle with fine-grained \emph{local details}---text, logos, thin structures---motivating a dedicated \emph{local refiner} for region-level detail correction.

\noindent\textbf{Image Editing Models.}
With increasingly capable generators, \emph{image editing} has gained growing attention~\cite{Nexus-Gen,wang2025gptimageedit15mmillionscalegptgeneratedimage,ye2025imgedit,liu2025step1x,cao_2023_masactrl,brooks2023instructpix2pix,seededit2024,xie2025reconstruction}. FLUX Kontext~\cite{labs2025fluxkontext} extends the text-only FLUX.1-dev~\cite{flux} by incorporating image inputs for editing. OmniGen2~\cite{wu2025omnigen2} uses modality-separated decoding with non-shared parameters and a decoupled image tokenizer, improving performance and consistency across generation, editing, and context-aware synthesis. BAGEL~\cite{bagel} proposes a Mixture-of-Transformers (MoT) design that couples an understanding model with a generator to better transfer instruction understanding. Qwen-Edit~\cite{wu2025qwen} encodes the input image with a VLM and injects its last-layer hidden states into a generative DiT, while also using a VAE to provide fine-detail context. Nevertheless, existing editing models largely focus on coarse-grained manipulations and often struggle with reliable \emph{fine-grained local refinement}, motivating \textsc{RefineAnything} for region-specific detail enhancement with strict background preservation.

\section{Method}

\subsection{Architecture}
We propose \methodname{} for \emph{localized refinement}. Given an input image $I$, an optional reference image $I^{\mathrm{ref}}$, a user-provided scribble mask $M$ indicating the edit region, and a text instruction $y$, our goal is to refine the specified region while preserving the rest of the image.

As shown in Fig.~\ref{fig:arch}, our overall framework builds on Qwen-Image~\cite{wu2025qwen} and consists of three components: (i) a frozen multimodal encoder (Qwen2.5-VL~\cite{Qwen2.5-VL}) that produces refinement-guiding conditioning tokens; (ii) a VAE that maps images to a latent space, providing fine-grained visual context; and (iii) a diffusion backbone built from MMDiT blocks that denoises a target latent under both multimodal and latent conditioning.

\textbf{High-level multimodal context (VLM).}
We encode the input (and optional reference) image, the region cue, and the instruction into multimodal conditioning tokens. Let $E_{\phi}(\cdot)$ denote the frozen Qwen2.5-VL encoder, then
\begin{equation}
\small
\mathbf{c} = E_{\phi}\big(I,\ I^{\mathrm{ref}},\ M,\ y\big), \qquad 
\mathbf{c}\in \mathbb{R}^{L\times d},
\label{eq:conditioning_tokens}
\end{equation}
where $L$ is the token length and $d$ is the feature dimension. These tokens provide high-level guidance (e.g., semantics and instruction intent) to the denoiser via joint-attention~\cite{zhou2025dreamrenderer,sd3,li2024hunyuandit,yang2024cogvideox,wu2025qwen}.

\textbf{Low-level visual context (VAE latents).}
We encode the input and optional reference images into VAE latents as low-level fine-grained visual conditioning:
\begin{equation}
\small
\mathbf{z}^{I} = \mathrm{Enc}_{\psi}(I), \qquad \mathbf{z}^{\mathrm{ref}} = \mathrm{Enc}_{\psi}(I^{\mathrm{ref}}) \in \mathbb{R}^{C\times H\times W},
\end{equation}
where $I^{\mathrm{ref}}$ is omitted if unavailable. These latents serve as additional conditioning branches (alongside the multimodal tokens $\mathbf{c}$ in Eq.~\ref{eq:conditioning_tokens}). We pack them with the noisy target latent $\mathbf{z}_t$ into patch token sequences and concatenate along the sequence dimension before feeding them into the MMDiT backbone.

\textbf{Denoising backbone (Qwen-Image).}
We adopt the MMDiT denoiser from Qwen-Image~\cite{wu2025qwen}. It iteratively removes noise from the target latent $\mathbf{z}_t$ conditioned on both the multimodal tokens $\mathbf{c}$ and the VAE latent branches.

\textbf{Inference.}
At inference, given $(I, I^{\mathrm{ref}}, M, y)$, we start from a noise latent $\mathbf{z}_T$ and iteratively denoise under the scheduler to obtain $\mathbf{z}_0$, which is decoded by the VAE decoder $\mathrm{Dec}_{\psi}$ into the output image $\widehat{I}$. Conditioning on $M$ steers refinement to the specified region while preserving the rest of the image.

\subsection{Focus-and-Refine}

\begin{figure*}[t]
  \centering
   \includegraphics[width=\linewidth]{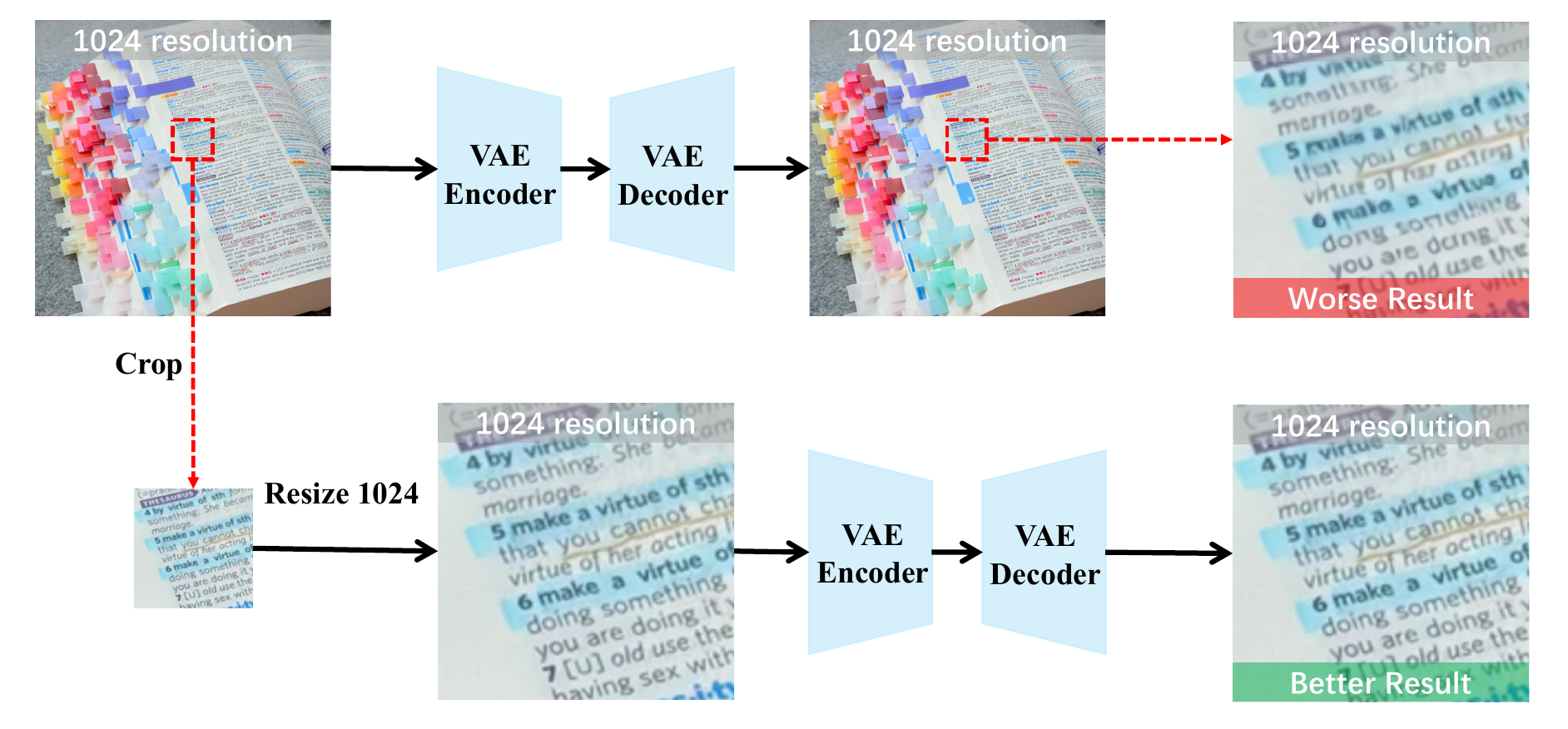}
\vspace{-5mm}
\caption{\textbf{Motivation for Focus-and-Refine.} We compare VAE reconstructing a local region (red box) from the full image versus first cropping the region and resizing it to original full image resolution before VAE encoding. \textit{Although the crop-and-resize step does not introduce new information, it substantially improves the reconstruction quality within the target region.} This observation suggests that, under a fixed input resolution, directing the model to focus on the local area rather than the entire image leads to better detail recovery for region-specific refinement. }
\label{fig:why_focus}
\vspace{-3mm}

\end{figure*}

\noindent\textbf{Motivation.}
Under a fixed input pixel budget (e.g., on the order of $1024\times1024$ pixels for VAE-based pipelines), local refinement is inherently challenging: the model receives only a limited amount of \emph{effective pixel information} about the fine structures to be repaired, since subtle details (e.g., thin strokes) may correspond to only a small number of pixels in the resized input. A natural question is whether we should process the \emph{entire} image under the same pixel budget, or instead focus the resolution budget on the region of interest.

Surprisingly, our experiments reveal a counter-intuitive phenomenon (Fig.~\ref{fig:why_focus}): \emph{although cropping the target region and resizing it to the same fixed resolution does not introduce any new information, it substantially improves reconstruction quality within the region.} In other words, simply re-parameterizing the input by zooming into the region—without changing the model, training data, or compute—already leads to sharper text strokes and cleaner local structures. This suggests that, for region-specific refinement, what limits quality is often not the \emph{availability} of information, but whether the model is forced to allocate its fixed-resolution capacity and attention to the right place.
This observation motivates our \emph{Focus-and-Refine} design

\begin{figure*}[t]
  \centering
   \includegraphics[width=\linewidth]{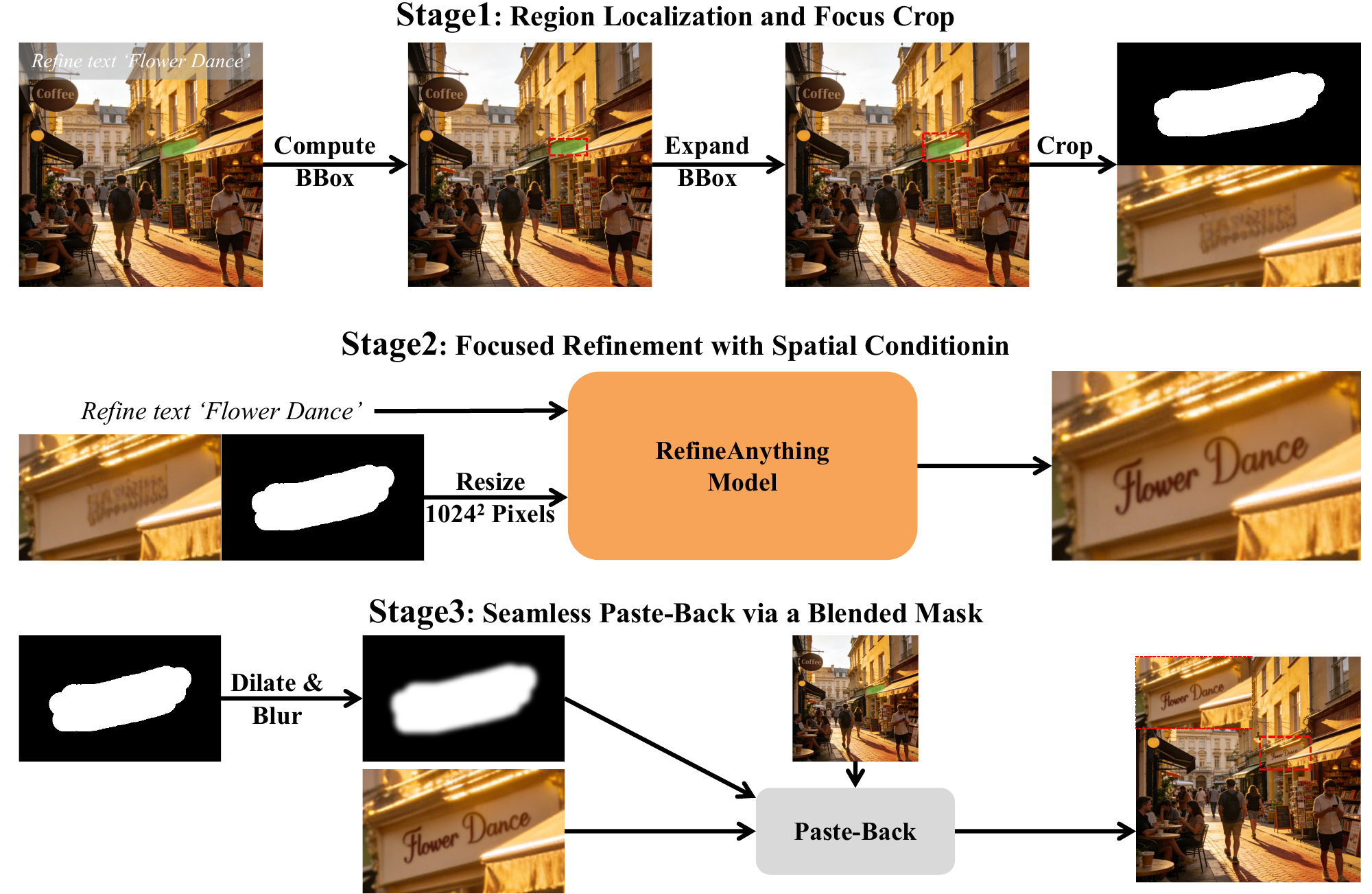}
\vspace{-5mm}
\caption{\textbf{Overview of Focus-and-Refine Method.} }
\label{fig:focus_and_paste}
\vspace{-3mm}

\end{figure*}
\noindent\textbf{Method.}
Given an input image $I\in\mathbb{R}^{H\times W\times 3}$, an optional reference image $I^{\mathrm{ref}}$, a text instruction $y$, and a scribble mask $M\in\{0,1\}^{H\times W}$, our goal is to generate a refined image $\widehat{I}$ such that the edit is localized to the region while the rest of the content is preserved.
As shown in Fig.~\ref{fig:focus_and_paste}, Focus-and-Refine consists of three steps: \emph{(i) region localization}, \emph{(ii) focused generation}, and \emph{(iii) seamless paste-back}.

\noindent\emph{(i) Region localization and focus crop.}
We first compute a tight bounding box around the scribble mask (or directly use the user-provided box),
\begin{equation}
\small
B = \mathrm{BBox}(M) = (x_1,y_1,x_2,y_2),
\label{eq:bbox_from_mask}
\end{equation}
and expand it with a margin $m$ to obtain the focus crop box
\begin{equation}
\small
C = \mathrm{Expand}(B,m)
\label{eq:expand_bbox}
\end{equation}
clipped to the image boundary. We then crop and resize the input (and the corresponding mask) to obtain the focused view:
\begin{equation}
\small
I_c = \mathrm{Crop}(I,C), \qquad M_c = \mathrm{Crop}(M,C).
\label{eq:crop_view}
\end{equation}
The margin $m$ provides local context (e.g., surrounding texture and illumination) while still concentrating most of the fixed-resolution budget on the target region.

\noindent\emph{(ii) Focused generation with spatial conditioning.}
On the cropped view, we use the cropped scribble mask $M_c$ as the spatial cue and perform conditional generation on a multi-image input:
\begin{equation}
\small
\mathcal{X} = \big\{ I_c,\ I^{\mathrm{ref}},\ M_c \big\},
\label{eq:multi_image_input}
\end{equation}
where $I^{\mathrm{ref}}$ is omitted if unavailable. The model then produces a refined crop
\begin{equation}
\small
\widetilde{I}_c = \mathcal{G}(\mathcal{X}, y),
\label{eq:refined_crop}
\end{equation}
where $\mathcal{G}$ denotes our RefineAnything Model (Fig.~\ref{fig:arch}).

\noindent\emph{(iii) Seamless paste-back via blended mask.}
Directly replacing the cropped area can introduce visible seams at the crop boundary. We therefore paste the refined result back using a softened version of the cropped mask $M_c$. Specifically, we apply morphological dilation and Gaussian smoothing to obtain a blended mask:
\begin{equation}
\small
\widetilde{M}_c = \mathrm{Blur}\big(\mathrm{Dilate}(M_c; r), k\big),
\label{eq:blend_mask}
\end{equation}
where $r$ is the dilate kernel size and $k$ is the blur kernel. We then composite the refined crop with the original crop:
\begin{equation}
\widehat{I}_c = \widetilde{M}_c \odot \widetilde{I}_c + (1-\widetilde{M}_c)\odot I_c,
\label{eq:composite_crop}
\end{equation}
with element-wise multiplication $\odot$. Finally, we resize and paste $\widehat{I}_c$ back to the full canvas at location $C$ to obtain the output image $\widehat{I}$. This design yields high-quality local refinement while maintaining global consistency, and the blended mask effectively suppresses boundary artifacts.

\subsection{Boundary Consistency Loss.}
To improve paste-back naturalness, we upweight supervision near the edit boundary during training.
We define a boundary band
\begin{equation}
\small
B_c=\mathrm{Dilate}(M_c;r_{\text{out}})-\mathrm{Erode}(M_c;r_{\text{in}}),
\label{eq:boundary_band}
\end{equation}
Following Qwen-Image~\cite{wu2025qwen}, we adopt the flow-matching denoising objective on the focused crop in latent space. Let $\mathbf{z}_0$ denote the latent of the target crop, sample $\mathbf{z}_1\sim\mathcal{N}(0,\mathbf{I})$ and $t\in[0,1]$, and construct $\mathbf{z}_t=t\mathbf{z}_0+(1-t)\mathbf{z}_1$ with target velocity $\mathbf{v}_t=\mathbf{z}_0-\mathbf{z}_1$. Conditioning on the multimodal tokens $\mathbf{c}$ in Eq.~\ref{eq:conditioning_tokens} and the VAE latent branches $\mathbf{z}^{I}$ (and $\mathbf{z}^{\mathrm{ref}}$ if available), the model predicts $\mathbf{v}_{\theta}(\mathbf{z}_t,t,\mathbf{c},\mathbf{z}^{I},\mathbf{z}^{\mathrm{ref}})$, yielding a per-location base loss map $\ell_{\text{base}}=\left\|\mathbf{v}_{\theta}(\mathbf{z}_t,t,\mathbf{c},\mathbf{z}^{I},\mathbf{z}^{\mathrm{ref}})-\mathbf{v}_t\right\|_2^2$ (summed over channels). We resize $B_c$ to match the spatial resolution of $\ell_{\text{base}}$ and define the boundary-weighted objective as
\begin{equation}
\small
\mathcal{L}_{\text{boundary}}
=\mathbb{E}\left[\left\|\ell_{\text{base}}\odot\left(1+\alpha B_c\right)\right\|_{1}\right].
\label{eq:boundary_weighted_loss}
\end{equation}

\subsection{Implementation Details}
\noindent\textbf{Training.} We fine-tune Qwen-Image-Edit~\cite{wu2025qwen} (2509 version) with LoRA~\cite{hu2022lora} on attention projections only (to\_q, to\_k, to\_v, to\_out.0): rank $256$, lora\_alpha $256$; only LoRA parameters are optimized. We use AdamW~\cite{adam} (lr $2\times10^{-4}$, $\beta_1$ $0.9$, $\beta_2$ $0.999$, weight decay $0.01$, $\epsilon$ $10^{-8}$) with a constant schedule, BF16, batch size $8$, and train for $20$K steps.
\textbf{Focus-and-Refine.} Crop margin $m=64$; paste-back mask uses Eq.~\ref{eq:blend_mask} with dilation kernel size $r=7$ and Gaussian blur kernel size $k=11$; boundary band uses Eq.~\ref{eq:boundary_band} with dilation/erosion kernel sizes $r_{\text{out}}=r_{\text{in}}=16$; boundary weighting uses Eq.~\ref{eq:boundary_weighted_loss} with $\alpha=9$.


\section{Refine-30K Dataset}
\label{sec:refine30kdata}
\begin{figure*}[t]
  \centering
   \includegraphics[width=\linewidth]{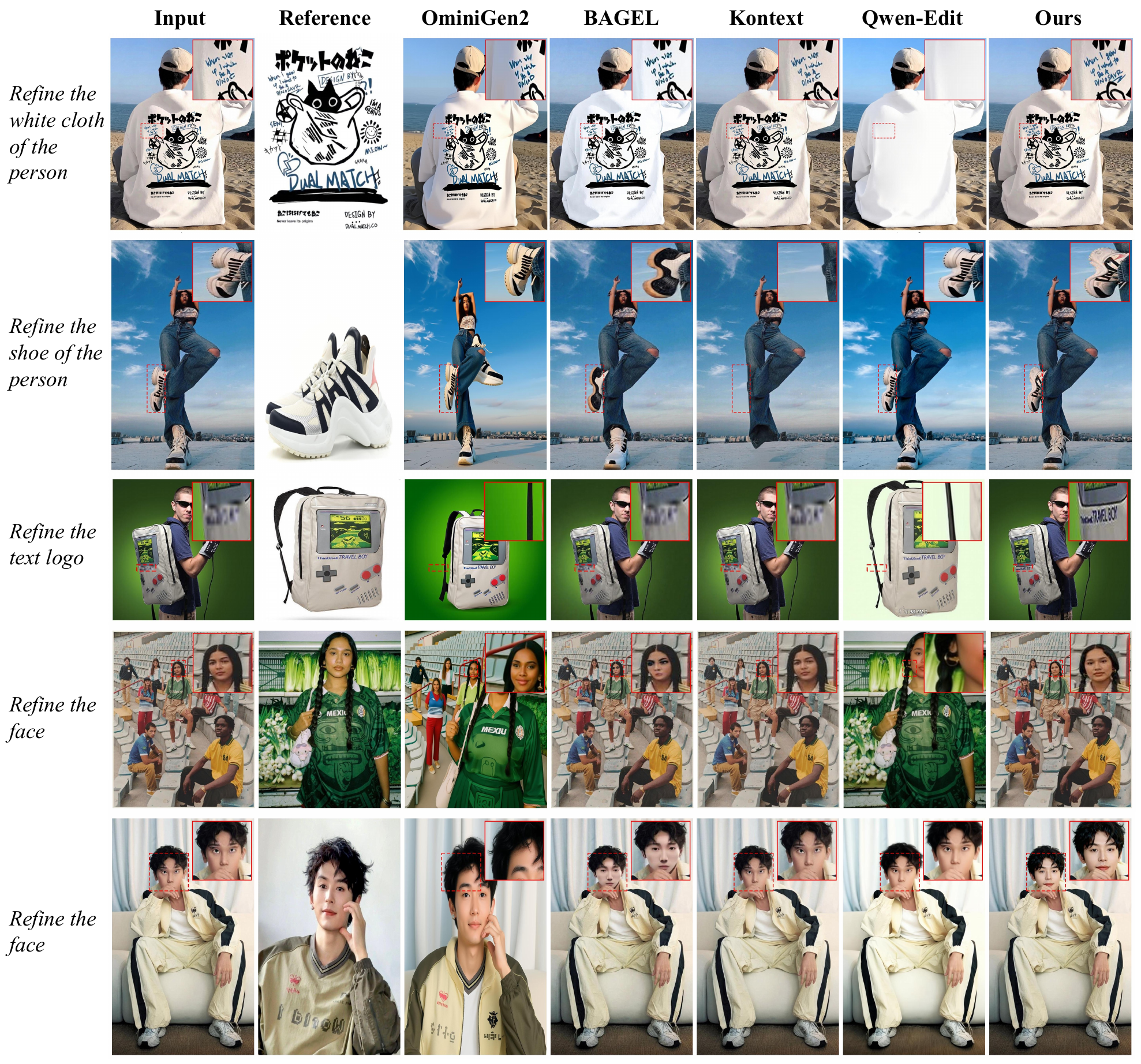}
\vspace{-5mm}
\caption{\textbf{Qualitative Result on Reference-Based Refinement.} }
\label{fig:qualitative_result}
\vspace{-3mm}

\end{figure*}

We collect \textbf{Refine-30K}, a dataset of 30K samples for training our \textsc{RefineAnything} model. Refine-30K consists of two subsets. The first subset contains \textbf{20K reference-based refine pairs}: as illustrated in Fig.~\ref{fig:qualitative_result}, the model is provided with both a refinement instruction and a reference image, and is expected to refine the input while following the visual style/appearance cues from the reference. The remaining \textbf{10K reference-free refine samples} are instruction-only: as shown in Fig.~\ref{fig:qualitative_result_free}, users provide only the refinement text to specify how the input should be refined.

\begin{figure*}[t]
  \centering
   \includegraphics[width=\linewidth]{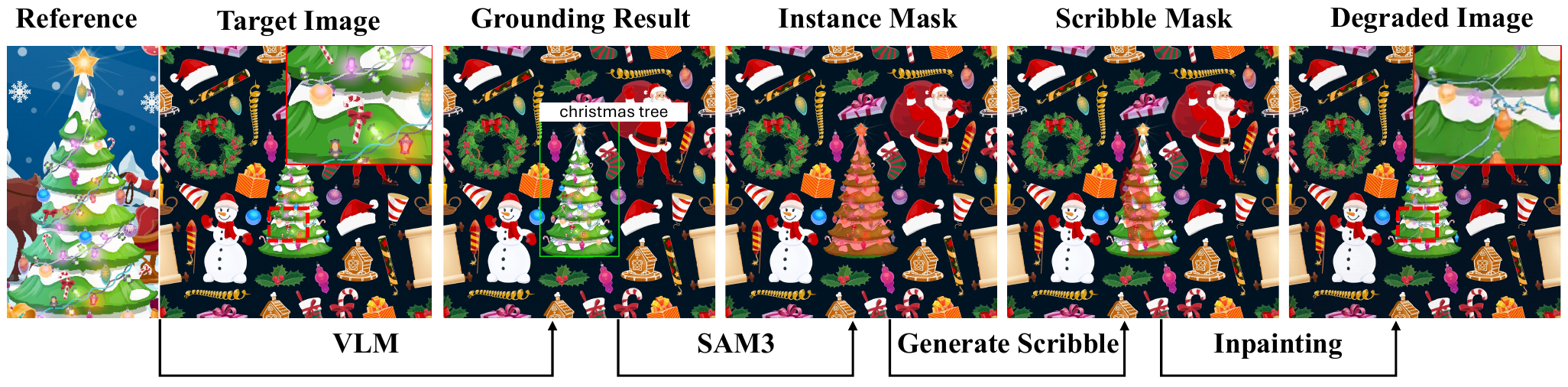}
\vspace{-5mm}
\caption{\textbf{Overview of Reference-Based Refine Data Construction Pipeline.} }
\label{fig:refine_base_process}
\vspace{-3mm}

\end{figure*}

\subsection{Reference-Based Refine Data}
We build the \textbf{reference-based} subset by converting each collected image pair into a supervised \emph{refinement} sample (Fig.~\ref{fig:refine_base_process}).
Each pair consists of a reference image $I^{\mathrm{ref}}$ and a target image $I^{\star}$, where $I^{\star}$ contains the main subject depicted in $I^{\mathrm{ref}}$.
Our pipeline produces a degraded \emph{input} image $I$ to be refined, the corresponding \emph{ground-truth} target $I^{\star}$, a spatial cue mask $M$, and a text instruction $y$ specifying the refinement goal. We construct each sample in four steps:

\noindent\emph{(i) Cross-image grounding.}
Given $(I^{\mathrm{ref}}, I^{\star})$, we apply a visual-language model (Gemini3~\cite{google2025gemini3promodelcard}) to identify the single most salient subject in $I^{\mathrm{ref}}$, verify that the same subject appears in $I^{\star}$, and localize it in $I^{\star}$ with a bounding box $B$.
To ensure high precision, we enforce strict subject-consistency checks and keep only pairs for which the VLM confidently confirms a match and outputs a valid box.

\noindent\emph{(ii) Mask generation with SAM.}
The bounding box may still include background clutter. We therefore refine localization by segmenting the subject region in $I^{\star}$.
Specifically, we run SAM (SAM3~\cite{carion2025sam3}) on the target image, conditioned on the VLM box and a short textual description, and obtain an object mask $M_{\mathrm{obj}}$.
We restrict to a single-instance mask to avoid ambiguous segmentations.

\noindent\emph{(iii) Scribble degradation via inpainting.}
To synthesize challenging refine inputs, we generate local artifacts within the localized subject region.
We first sample random scribble strokes and constrain them to lie inside a dilated version of $M_{\mathrm{obj}}$, yielding the final inpainting mask $M$.
We then inpaint the target image to obtain a degraded image $\widetilde{I}$:
\begin{equation}
\widetilde{I} = \mathrm{Inpaint}(I^{\star}, M).
\end{equation}
This step introduces realistic local corruptions while keeping the degradation spatially controlled, and we apply a light paste-back blending to ensure the final input differs from $I^{\star}$ only within the edited region.

\noindent\emph{(iv) Instruction and outputs.}
For each sample, we store $(I, I^{\mathrm{ref}}, I^{\star}, M, y)$.
The instruction $y$ is derived from the VLM description and explicitly refers to the localized region to align with our region-conditioned refinement setting.

\subsection{Reference-Free Refine Data}
We construct a \textbf{reference-free} subset from single images, using only a refinement instruction and a spatial cue (no external reference). We synthesize a degraded input while keeping the original as ground truth, and employ a VLM to filter implausible or unrecognizable degradations to keep the task well-defined.
We construct each sample in four steps:

\noindent\emph{(i) Salient object localization.}
Given a single image $I^{\star}$, we first apply a VLM (Gemini3~\cite{google2025gemini3promodelcard}) to detect salient objects and produce a set of candidate bounding boxes $\{B_i\}$ along with short textual descriptions. We then randomly sample one object $B$ to diversify the edited regions across categories and scales.

\noindent\emph{(ii) Masking and degradation.}
We then follow the same segmentation and scribble-based inpainting degradation pipeline as in the reference-based subset: we obtain an object mask $M_{\mathrm{obj}}$ using SAM3~\cite{carion2025sam3}, sample a scribble mask $M$ inside a dilated $M_{\mathrm{obj}}$, and synthesize a degraded input $I$ from $I^{\star}$ via inpainting with a light paste-back blending so that $I$ and $I^{\star}$ differ only in the intended region.

\noindent\emph{(iii) VLM-based defect validation.}
Not all synthetic degradations lead to meaningful refinement tasks. We therefore employ a VLM to judge whether the degraded image $I$ exhibits noticeable defects (e.g., artifacts, missing structures, or unnatural textures) and whether the degradation is logically plausible given the scene. We discard samples that are judged as (a) having no obvious defect or (b) being semantically inconsistent/ill-posed, which improves data quality and stabilizes training.

\noindent\emph{(iv) Instruction and outputs.}
Each sample is stored as $(I, I^{\star}, M, y)$, where $y$ is a reference-free refinement instruction generated from the VLM description of the selected object/region (e.g., ``Refine \{object\} in the masked region''). This subset complements reference-based data by teaching the model to follow text-only refinement cues while maintaining strict background consistency.

\section{Experiment}
\subsection{Benchmarks}
To evaluate the image refinement capabilities of our model, we construct \textbf{RefineEval}. RefineEval includes two settings: \textbf{Reference-Based Image Refinement} and \textbf{Reference-Free Image Refinement}. The former focuses on identity-sensitive content such as specific logo text, products, and person IDs, while the latter covers common structures including human bodies, generic objects, faces, and text.
Each RefineEval \emph{case} provides a clean target image, a localized edit region, and a refinement instruction (and additionally a reference image in the reference-based setting). We curate 67 cases from open-source websites and manually annotate the regions to be degraded/refined. Following the data construction protocol in Sec.~\ref{sec:refine30kdata}, we synthesize degraded inputs via inpainting within the annotated regions, using Flux-fill \cite{flux}, SDXL \cite{sdxl}, and Qwen-Edit \cite{wu2025qwen} to cover varying degradation patterns. For each inpainting method, we generate candidates with 5 randomly sampled scribble masks across 3 different seeds and manually select 2 representative degraded images for evaluation. This results in 402 degraded inputs in total (67 cases $\times$ 3 methods $\times$ 2 images), including 31 reference-based cases and 36 reference-free cases.

\begin{table}[t]
\centering
\caption{\textbf{Evaluation on Reference-Based Image Refinement.}}
\setlength{\tabcolsep}{0.5pt}  
\renewcommand{\arraystretch}{0.95}  
\small
\begin{tabular}{@{}l
                cccccccc  
                p{1.2cm}@{}}  
\hline\hline
\textbf{Method} 
& \textbf{MSE$\downarrow$ }
& \textbf{LP$\downarrow$ }
& \textbf{VGG$\downarrow$ }
& \textbf{DINO$\uparrow$ }
& \textbf{CLIP$\uparrow$ }
& \textbf{SSIM$\uparrow$ }
& \textbf{MSE$_{\text{bg}}$$\downarrow$ }
& \textbf{LP$_{\text{bg}}$$\downarrow$ }
& \multicolumn{1}{c@{}}{\textbf{SSIM$_{\text{bg}}$$\uparrow$}} \\  
\hline

Gemini2.5 & 0.049 & 0.250 & 0.592 & 0.717 & 0.817 & 0.423 & 0.201 & 0.103  & 0.7662 \\
Gemini3   & 0.031 & 0.178 & 0.431 & 0.771 & 0.855 & 0.510 & 0.029 & 0.052 & 0.9061 \\
GPT4o     & 0.083 & 0.370 & 0.918 & 0.620 & 0.801 & 0.302 & 0.815 & 0.309 & 0.6001 \\

\hline
OmniGen2  & 0.155 & 0.602 & 1.691 & 0.384 & 0.717 & 0.219 & 2.094 & 0.624 & 0.4300 \\
BAGEL     & 0.045 & 0.253 & 0.611 & 0.682 & 0.803 & 0.494 & 0.033 & 0.046 & 0.9360 \\
Kontext   & 0.040 & 0.264 & 0.540 & 0.685 & 0.785 & 0.538 & 0.011 & 0.019 & 0.9660 \\
Qwen-Edit & 0.049 & 0.287 & 0.676 & 0.675 & 0.807 & 0.436 & 0.454 & 0.148 & 0.7530 \\
\hline
\rowcolor{blue!15}
\textbf{Ours} 
& \textbf{0.020} & \textbf{0.155} & \textbf{0.401}
& \textbf{0.793} & \textbf{0.885} & \textbf{0.591}
&  \textbf{0.000} & \textbf{0.000} & \textbf{0.9997} \\
\hline\hline
\end{tabular}
\parbox{\linewidth}{\vspace{2pt}\footnotesize 
$\downarrow$: Smaller is better, $\uparrow$: Larger is better. Gemini2.5 represents Gemini2.5 flash image, Gemini3 represents Gemini3-pro. LP stands for the LPIPS metric, and DINO stands for the DINOv2Large metric.}
\label{tab:reference_base}
\end{table}

\subsection{Evaluation Metrics}

\noindent \textbf{Reference-Based Image Refinement.}
When a reference image is provided, we evaluate (i) edited-region fidelity and (ii) background preservation. For the edited region, we compare the refined image with the ground-truth (GT) image using MSE, SSIM, LPIPS, VGG, and feature similarities via DINO and CLIP; for the background, we compare the refined image with the input image using $\text{MSE}_{bg}$, $\text{LPIPS}_{bg}$, and $\text{SSIM}_{bg}$. Foreground/background regions are defined by the object bounding box annotations in the benchmark. We use dino-v2 large \cite{oquab2023dinov2} for DINO and clip-vit-large-patch14-336 \cite{chen2022altclip} for CLIP.


\begin{figure*}[t]
  \centering
   \includegraphics[width=\linewidth]{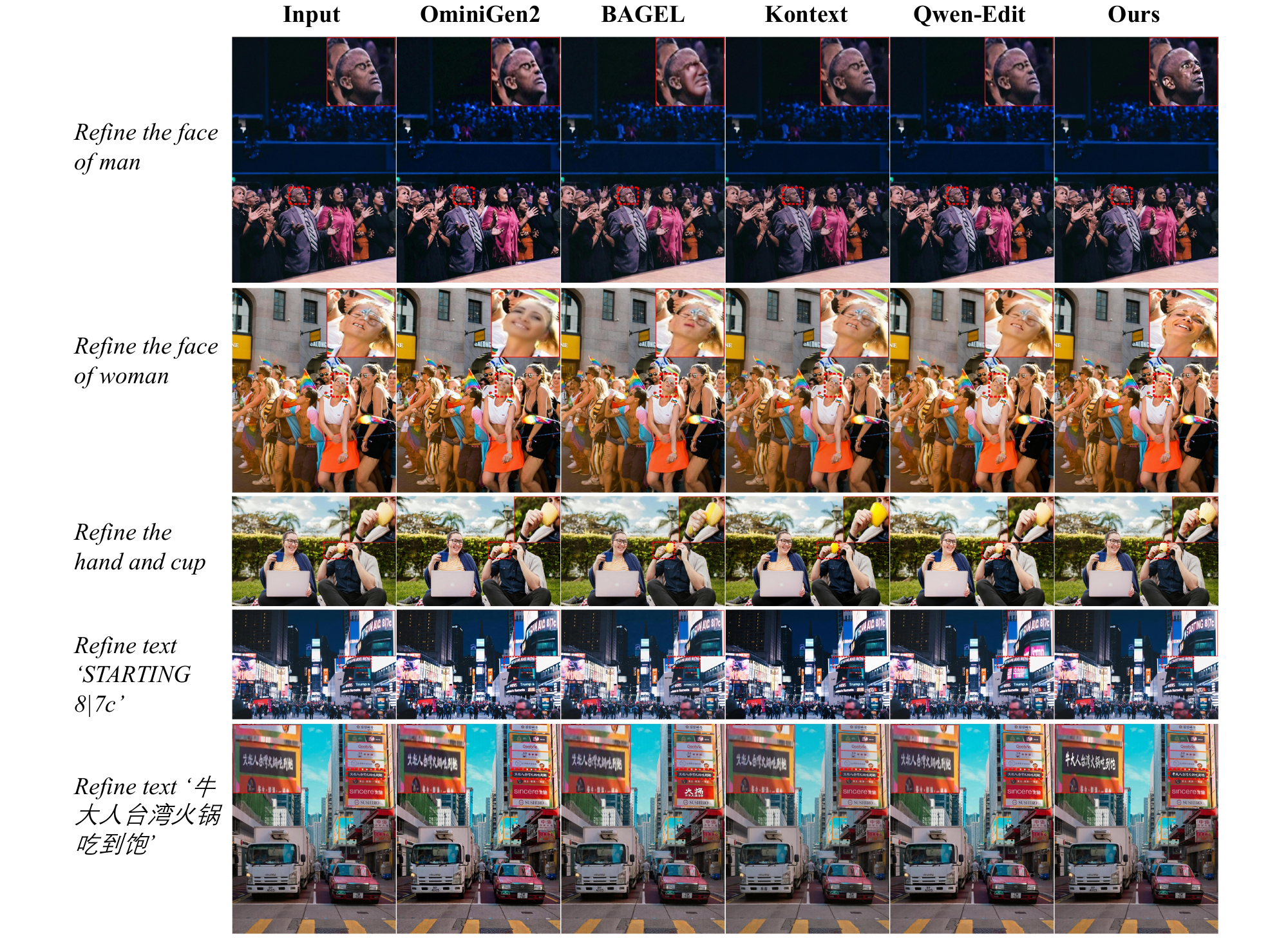}
\vspace{-5mm}
\caption{\textbf{Qualitative Result on Reference-Free Refinement.} }
\label{fig:qualitative_result_free}
\vspace{-3mm}

\end{figure*}

\noindent \textbf{Reference-Free Image Refinement.}
In the absence of a reference image, refinement is inherently open-ended. We therefore adopt a VLM-based evaluator (Gemini2.5-Pro) and score the expanded foreground crop for each case on five dimensions: visual quality (\textbf{VQ}), naturalness (\textbf{Nat.}), aesthetics (\textbf{Aes.}), fine-detail fidelity (\textbf{Det.}), and instruction faithfulness (\textbf{Faith.}). Scores are in $[1,5]$ with decimals allowed (higher is better); prompts are provided in the appendix.
\begin{table}[t]
\centering
\caption{\textbf{Evaluation on the Reference-Free Image Refinement.}}
\setlength{\tabcolsep}{4pt}  
\renewcommand{\arraystretch}{0.95}  
\begin{tabular}{@{}lccccc@{}}  
\hline\hline  
\textbf{Method} 
& \textbf{VQ$\uparrow$} 
& \textbf{Nat.$\uparrow$} 
& \textbf{Aes.$\uparrow$} 
& \textbf{Det.$\uparrow$} 
& \textbf{Faith.$\uparrow$} \\  
\hline  
OmniGen2  & 2.501 & 2.500 & 2.461 & 2.348 & 2.586 \\
BAGEL     & 3.018 & 3.000 & 2.959 & 2.851 & 3.135  \\
kontext   & 1.716 & 2.114 & 1.982 & 1.690 & 1.750  \\
Qwen-Edit & 3.081 & 3.110 & 3.105 & 2.975 & 3.214  \\
\hline
\rowcolor{blue!15}
\textbf{Ours} & \textbf{3.806} & \textbf{3.868} & \textbf{3.876} & \textbf{3.720} & \textbf{3.644} \\
\hline\hline  
\end{tabular}
\label{tab:subjective_evaluation}
\end{table}

\begin{table}[t]
\centering
\caption{\textbf{Ablation on Focus-and-Refine and Boundary Consistency Loss.}}
\setlength{\tabcolsep}{0.5pt}  
\renewcommand{\arraystretch}{0.95}  
\small
\begin{tabular}{@{}l
                cccccccc  
                p{1.2cm}@{}}  
\hline\hline
\textbf{Method} 
& \textbf{MSE$\downarrow$ }
& \textbf{LP$\downarrow$ }
& \textbf{VGG$\downarrow$ }
& \textbf{DINO$\uparrow$ }
& \textbf{CLIP$\uparrow$ }
& \textbf{SSIM$\uparrow$ }
& \textbf{MSE$_{\text{bg}}$$\downarrow$ }
& \textbf{LP$_{\text{bg}}$$\downarrow$ }
& \multicolumn{1}{c@{}}{\textbf{SSIM$_{\text{bg}}$$\uparrow$}} \\  
\hline

w/o focus  & 0.021 & 0.177 & 0.449 & 0.779 & 0.869 & 0.578 & 0.005 & 0.022 & 0.9601 \\
w/o loss  & 0.023 & 0.191 & 0.482 & 0.736 & 0.858 & 0.563 & 0.000 & 0.000 & 0.9997 \\
\hline
\rowcolor{blue!15}
\textbf{Ours} 
& \textbf{0.020} & \textbf{0.155} & \textbf{0.401}
& \textbf{0.793} & \textbf{0.885} & \textbf{0.591}
&  \textbf{0.000} & \textbf{0.000} & \textbf{0.9997} \\
\hline\hline
\end{tabular}
\label{tab:ablation}
\end{table}

\subsection{Baselines}
We compare our method with several representative open-source and closed-source approaches for image editing and instruction-based generation, including  GPT4o~\cite{gpt-4o}, Gemini 3-pro-image-preview~\cite{google2025gemini3promodelcard}, Gemini 2.5-flash-image~\cite{google2025gemini25flashmodelcard}, Qwen-Image-Edit~\cite{wu2025qwen}, BAGEL~\cite{bagel}, OmniGen2~\cite{wu2025omnigen2}, and Kontext~\cite{labs2025fluxkontext}. More details are provided in the supplementary material.


\subsection{Quantitative Results}
Tab.~\ref{tab:reference_base} shows that our method achieves the best overall performance on reference-based refinement, jointly improving \emph{edited-region fidelity} and \emph{background preservation}. Compared to the best open-source baseline (Kontext), we reduce MSE by 0.020 (50\%), LPIPS by 0.109 (41\%), and VGG by 0.139 (26\%), and improve DINO and CLIP by +0.108 and +0.100, respectively. Meanwhile, we deliver near-perfect background consistency ($\mathrm{MSE}_{bg}=0.000$, $\mathrm{LP}_{bg}=0.000$, $\mathrm{SSIM}_{bg}=0.9997$), eliminating background drift (e.g., Kontext: $\mathrm{MSE}_{bg}=0.011$; Qwen-Edit: $\mathrm{MSE}_{bg}=0.454$).
For reference-free refinement, Tab.~\ref{tab:subjective_evaluation} reports MLLM-based subjective scores across five dimensions (VQ, Nat., Aes., Det., and Faith.). Our method ranks first on all criteria, surpassing the strongest open-source baseline (Qwen-Edit) by +0.725, +0.758, +0.771, +0.745, and +0.430 on VQ, Nat., Aes., Det., and Faith., respectively, indicating more natural, detailed, and instruction-faithful refinements even without a reference image.

\subsection{Qualitative Results}

Fig.~\ref{fig:qualitative_result} and Fig.~\ref{fig:qualitative_result_free} present a qualitative comparison between our method and state-of-the-art baselines on reference-based and reference-free refinement. Prior methods often suffer from poor background preservation, weak responsiveness to the instruction/reference, and limited ability to recover fine details. In contrast, empowered by our Focus-and-Refine strategy, our approach not only restores subtle details more effectively but also keeps the background strictly unchanged, substantially improving practicality and real-world usability.

\subsection{Ablation Study}
\begin{figure*}[t]
  \centering
   \includegraphics[width=\linewidth]{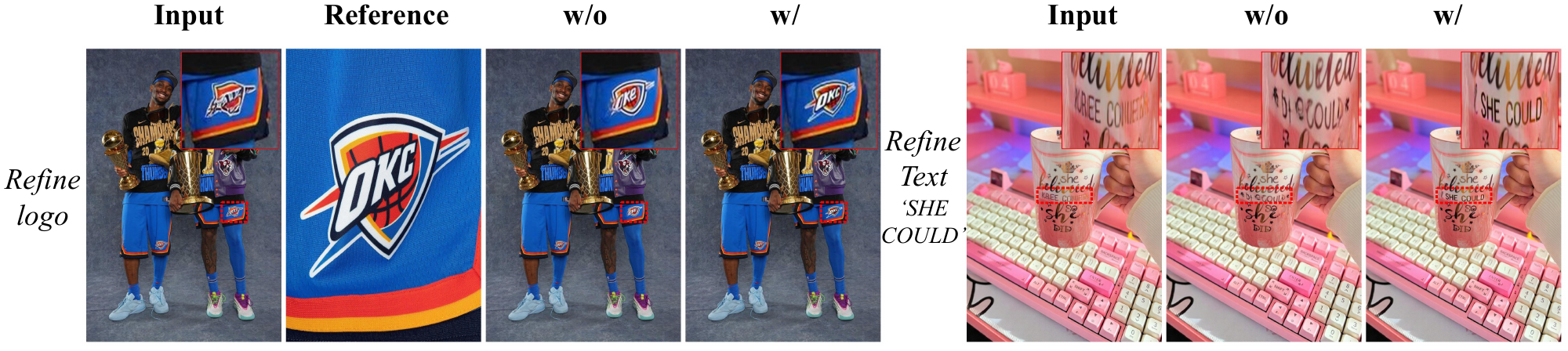}
\vspace{-5mm}
\caption{\textbf{Ablation of the Focus-and-Refine strategy.} }
\label{fig:vs_focus_and_refine}
\vspace{-3mm}

\end{figure*}
\begin{figure*}[t]
  \centering
   \includegraphics[width=\linewidth]{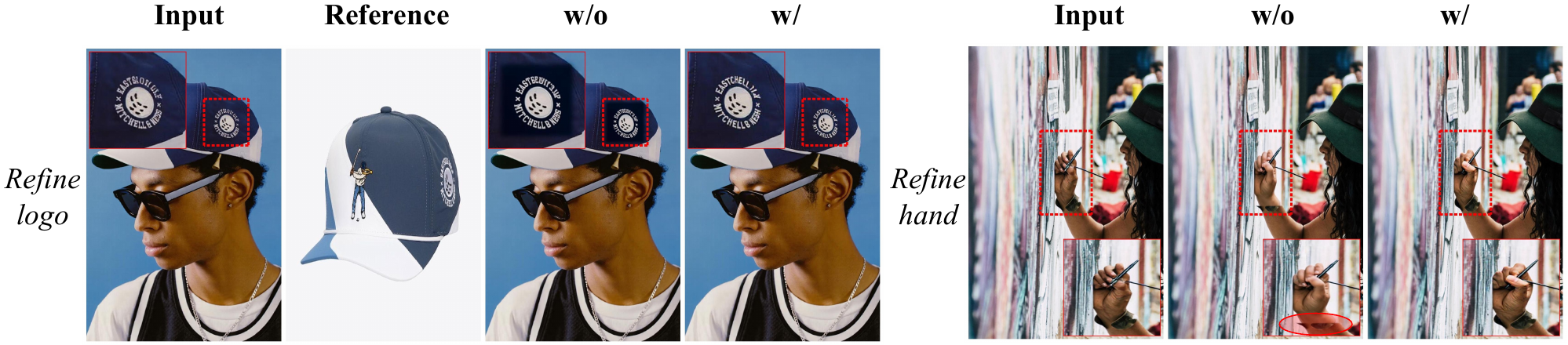}
\vspace{-5mm}
\caption{\textbf{Ablation of the Boundary Consistency Loss.} }
\label{fig:vs_loss}
\vspace{-3mm}

\end{figure*}

\noindent\textbf{Focus-and-Refine.} As shown in Fig.~\ref{fig:vs_focus_and_refine} and Tab.~\ref{tab:ablation}, removing the focusing step leads to weaker refinements, often leaving subtle errors unresolved and occasionally introducing artifacts. In contrast, Focus-and-Refine allocates the model’s capacity to the target region, producing sharper local details while keeping the surrounding background unchanged.

\noindent\textbf{Boundary Consistency Loss.} As shown in Fig.~\ref{fig:vs_loss} and Tab.~\ref{tab:ablation}, removing the Boundary Consistency Loss leads to poor coherence between the locally refined region and its surrounding context. This often manifests as visible seams, color inconsistencies, and structurally implausible stitching along object boundaries.



\section{Conclusion}
We introduced \textsc{RefineAnything}, the first framework tailored for \emph{region-specific image refinement}---improving fine-grained local details while keeping non-edited regions strictly unchanged. Motivated by the observation that crop-and-resize can significantly boost local reconstruction under a fixed input resolution, we proposed \emph{Focus-and-Refine}, which concentrates model capacity on the target region and then pastes the refined result back with a blended mask for seamless integration. To further enhance paste-back naturalness, we introduced a boundary-aware \emph{Boundary Consistency Loss} that encourages seam-consistent refinements during training. We also built \textbf{Refine-30K} and the \textbf{RefineEval} benchmark to support training and evaluation in both reference-based and reference-free settings. Extensive experiments demonstrate that our approach improves local detail fidelity and semantic alignment while achieving near-perfect background preservation. We hope \textsc{RefineAnything} and our datasets will facilitate future research on practical, high-precision refinement for real-world image generation and editing workflows.

\bibliographystyle{splncs04}
\bibliography{main}
\end{document}